\documentclass[journal]{IEEEtran}
\usepackage{amsmath}
\usepackage{algorithm}
\usepackage{algorithmic}
\usepackage{graphicx}
\usepackage{multirow}
\usepackage{tabularx}
\usepackage{xcolor}
\usepackage{array}
\usepackage{amssymb}
\usepackage{bbding}
\usepackage{subfigure}
\usepackage{diagbox}
\usepackage{rotating}
\usepackage{bm}
\usepackage{tikz} 
\usepackage{booktabs}
\usepackage{cite}
\usepackage{CJK}
\usepackage{overpic}
\usepackage[colorlinks,
            linkcolor=red,
            anchorcolor=blue,
            citecolor=green
            ]{hyperref}

\newcommand{\addFig}[1]{}
\newcommand{\addFigs}[1]{}

\usepackage{subfigure}
\newcommand{\etal}{\textit{et~al}.~}
\newcommand{\ie}{\textit{i}.\textit{e}.,~}
\newcommand{\eg}{\textit{e}.\textit{g}.,~}

\usepackage{balance}
\usepackage{cleveref}
\crefformat{section}{\S#2#1#3} 
\crefformat{subsection}{\S#2#1#3}
\crefformat{subsubsection}{\S#2#1#3}

\ifCLASSINFOpdf
\else
\fi

\begin{document}
%
\title{Lightweight Salient Object Detection in Optical Remote Sensing Images via Feature Correlation}
%
%
%

\author{Gongyang~Li,
	Zhi~Liu,~\IEEEmembership{Senior Member,~IEEE},
	Zhen~Bai,\\
	Weisi~Lin,~\IEEEmembership{Fellow,~IEEE},
        and~Haibin~Ling,~\IEEEmembership{Senior Member,~IEEE}
        
\thanks{Gongyang Li, Zhi Liu, and Zhen Bai are with Shanghai Institute for Advanced Communication and Data Science, Shanghai University, Shanghai 200444, China, and School of Communication and Information Engineering, Shanghai University, Shanghai 200444, China (email: ligongyang@shu.edu.cn; liuzhisjtu@163.com; bz536476@163.com).}
\thanks{Weisi Lin is with the School of Computer Science and Engineering, Nanyang Technological University, Singapore 639798 (e-mail: wslin@ntu.edu.sg).}
\thanks{Haibin Ling is with the Department of Computer Science, Stony Brook University, Stony Brook, NY 11794 USA (email: hling@cs.stonybrook.edu).}
\thanks{\textit{Corresponding author: Zhi Liu.}}
}

\markboth{IEEE TRANSACTIONS ON GEOSCIENCE AND REMOTE SENSING}%
{Shell \MakeLowercase{\textit{et al.}}: Bare Demo of IEEEtran.cls for IEEE Journals}

\maketitle

\begin{abstract}
Salient object detection in optical remote sensing images (ORSI-SOD) has been widely explored for understanding ORSIs.
However, previous methods focus mainly on improving the detection accuracy while neglecting the cost in memory and computation, which may hinder their real-world applications.
In this paper, we propose a novel lightweight ORSI-SOD solution, named \emph{CorrNet}, to address these issues.
In CorrNet, we first lighten the backbone (VGG-16) and build a lightweight subnet for feature extraction.
Then, following the coarse-to-fine strategy, we generate an initial coarse saliency map from high-level semantic features in a Correlation Module (CorrM).
The coarse saliency map serves as the location guidance for low-level features.
In CorrM, we mine the object location information between high-level semantic features through the cross-layer correlation operation.
Finally, based on low-level detailed features, we refine the coarse saliency map in the refinement subnet equipped with Dense Lightweight Refinement Blocks, and produce the final fine saliency map.
By reducing the parameters and computations of each component, CorrNet ends up having only 4.09M parameters and running with 21.09G FLOPs.
Experimental results on two public datasets demonstrate that our lightweight CorrNet achieves competitive or even better performance compared with 26 state-of-the-art methods (including 16 large CNN-based methods and 2 lightweight methods), and meanwhile enjoys the clear memory and run time efficiency.
The code and results of our method are available at https://github.com/MathLee/CorrNet.
\end{abstract}


\IEEEpeerreviewmaketitle

\section{Introduction}
\IEEEPARstart{S}{alient} object detection (SOD)~\cite{2019sodsurvey,WWG19Video,19CRMCO} focuses on extracting the visually distinctive objects or regions in a scene, and often serves as an important preprocessing step in computer vision.
It has been successfully applied in image retargeting~\cite{2012Retargeting}, image quality assessment~\cite{16SODIQA,19SGDNet}, and object segmentation~\cite{LGY2019,LGY2021PFOS}, \textit{etc}.
In recent decades, there have been many branches of SOD, such as SOD in natural scene images (NSI-SOD)~\cite{2019sodsurvey}, video SOD~\cite{WWG19Video}, RGB-D SOD~\cite{20ICNet,20CMWNet}, co-saliency detection~\cite{2021AGCNet}, SOD in optical remote sensing images (ORSI-SOD)~\cite{2019LVNet}, \textit{etc}.
In this paper, we are committed to an emerging topic in SOD, \ie ORSI-SOD.
Optical remote sensing images (ORSIs) are photographed by satellites and aerial sensors, and have three optical bands (\ie red, green and blue bands), which are the same as natural scene images (NSIs).
The scenes of ORSIs are completely different from NSIs and are very challenging.
ORSI-SOD can discover attractive objects, which is conducive to quickly analyzing and understanding ORSIs.

\begin{figure}[t!]
  \centering
  \footnotesize
  \begin{overpic}[width=0.99\columnwidth]{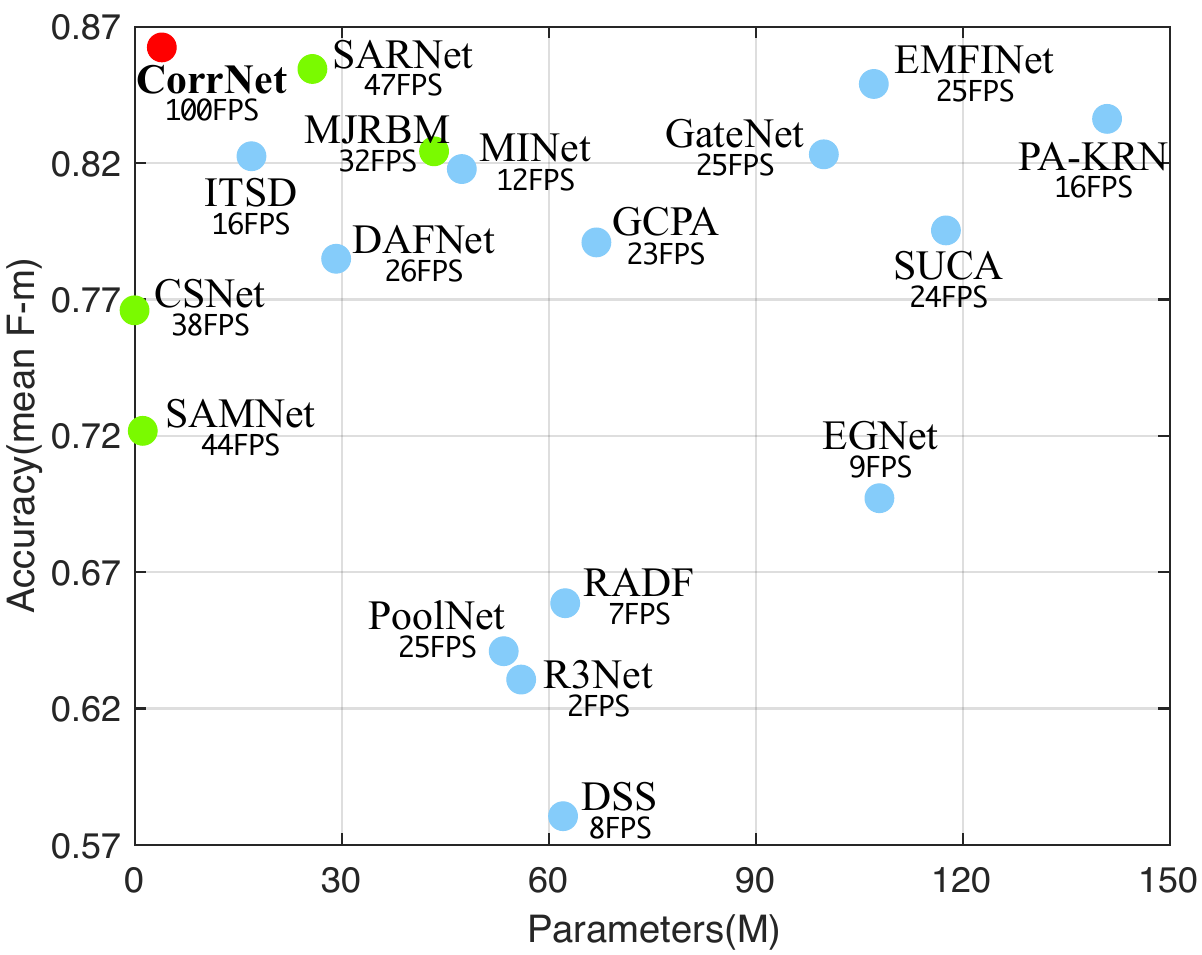}
  \end{overpic}
  \caption{Accuracy, parameters and inference speed comparisons of our CorrNet and other CNN-based methods on the EORSSD dataset~\cite{2021DAFNet}.
  \textcolor[rgb]{0.48 0.98 0}{\large{$\bullet$}} represents real-time methods, \textcolor[rgb]{0.52 0.80 0.98}{\large{$\bullet$}} represents non-real-time methods, and \textcolor{red}{\large{$\bullet$}} represents our CorrNet.
    }\label{fig:example}
\end{figure}

Early traditional NSI-SOD methods~\cite{2015SODBenchmark} mainly relied on hand-crafted features, which usually lead to unsatisfactory detection accuracy.
Recently, convolutional neural networks (CNNs)~\cite{1989CNN} have demonstrated powerful capabilities in computer vision, and greatly promoted the development of NSI-SOD algorithms~\cite{2019sodsurvey}, which often generate satisfactory saliency maps.
However, the improvement in detection accuracy often comes from more complicated network structures, which typically come with a large amount of parameters and increased computational complexity.
Since ORSIs and NSIs have gaps in the scene, it is not appropriate to directly migrate NSI-SOD to ORSIs, but most of the existing ORSI-SOD methods~\cite{2021DAFNet,2021EMFINet,2021MJRBM,2021SARNet} are affected by NSI-SOD methods.
Therefore, ORSI-SOD methods usually have a large computational consumption and memory burden, and are accompanied by limited inference speed.

In Fig.~\ref{fig:example}, we show the detection accuracy of recent NSI-SOD methods (PA-KRN~\cite{2021PAKRN}, SUCA~\cite{2021SUCA}, GateNet~\cite{2020GateNet}, EGNet~\cite{2019EGNet}) and ORSI-SOD methods (DAFNet~\cite{2021DAFNet}, EMFINet~\cite{2021EMFINet}, MJRBM~\cite{2021MJRBM}, SARNet~\cite{2021SARNet}) on an ORSI-SOD dataset, namely EORSSD~\cite{2021DAFNet}.
We also report their parameters and inference speed in Fig.~\ref{fig:example}.
Although these methods show good performance, their parameters are amazing and inference speeds are slow, for example, PA-KRN has 141.06M parameters with only 16 fps and EMFINet has 107.26M parameters with 25 fps.
And the performance of the two lightweight NSI-SOD methods (CSNet~\cite{2020CSNet} and SAMNet~\cite{2021SAMNet}) is slightly inferior.
Considering the application scenarios of ORSI-SOD, we believe that ORSI-SOD is in urgent need of a lightweight solution with fewer parameters, faster speed and good accuracy.

Inspired by the above observations, in this paper, we propose a novel lightweight solution for ORSI-SOD, namely \emph{CorrNet}, which is the first lightweight ORSI-SOD model as we know.
In CorrNet, we mainly realize the lightweight framework from two aspects: 1) lightening the backbone, and 2) designing lightweight modules.
For the backbone, previous methods~\cite{2021DAFNet,2021EMFINet,2021MJRBM,2021SARNet} usually adopt the pre-trained VGG~\cite{2014VGG16ICLR} or ResNet~\cite{2016ResNet} as the backbone, but such backbones suffer from a large number of parameters despite their powerful feature extraction capabilities.
To achieve a balance between the feature extraction capabilities and the amount of parameters, we modify the vanilla VGG~\cite{2014VGG16ICLR} and build a lightweight but powerful backbone for feature extraction.
For the lightweight modules, we use the depthwise separable convolution~\cite{MobileNet1,MobileNet2} instead of the regular one, which can reduce the parameters of regular convolution by about 90\%.
In this way, our CorrNet has only 4.09M parameters.

Moreover, to keep a good detection accuracy of CorrNet, we implement it following the coarse-to-fine strategy with two novel modules.
For the coarse part, we explore the object location information among two groups of high-level semantic features in the \emph{Correlation Module}, and obtain the initial \emph{coarse} saliency map.
Then, we refine it with other low-level detailed features in the refinement subnet, which consists of several \emph{Dense Lightweight Refinement Blocks}, and obtain the final \emph{fine} saliency map.
With all components working together, our CorrNet achieves excellent performance in accuracy (86.20\% in mean F-measure on the EORSSD dataset~\cite{2021DAFNet}), parameters (4.09M) and inference speed (100fps), as shown in Fig.~\ref{fig:example}.

Our main contributions are summarized as follows:
\begin{itemize}
\item We explore the lightweight framework of ORSI-SOD for the first time.
To this end, we propose a novel lightweight \emph{CorrNet} (only 4.09M parameters) that uses the \emph{coarse-to-fine} strategy.

\item We propose a \emph{Correlation Module} to explore the cross-layer correlation of high-level semantic context, generating an initial \emph{coarse} saliency map to low-level features for location guidance.

\item We propose a \emph{Dense Lightweight Refinement Block} to merge the enhanced feature embeddings and the refined features for finely sculpting salient objects, gradually producing the final \emph{fine} saliency map.

\item We evaluate the proposed CorrNet against 26 state-of-the-art methods on two ORSI-SOD datasets.
Experiments demonstrate that the proposed CorrNet achieves better or competitive performance compared with previously proposed large CNN-based methods.

\end{itemize}

We organize the rest of this paper as follows.
In Sec.~\ref{sec:related}, we review the related work of ORSI-SOD.
In Sec.~\ref{sec:OurMethod}, we elaborate our CorrNet.
In Sec.~\ref{sec:exp}, we conduct experiments and ablation studies.
Finally, in Sec.~\ref{sec:con}, we draw the conclusion.

\section{Related Work}
\label{sec:related}

\subsection{Lightweight Methods for NSI-SOD}
\label{sec:Light_SOD}
The lightweight NSI-SOD task is an emerging direction in NSI-SOD, which aims to propose a solution suitable for edge computing devices.
Gao~\etal\cite{2020CSNet} proposed an extremely lightweight network with only about 100K parameters and 95.3ms run-time on a single core i7-8700K CPU.
Liu~\etal\cite{2021SAMNet} constructed a stereoscopic attention mechanism based backbone for lightweight NSI-SOD.
In~\cite{2021HVPNet}, Liu~\etal imitated the primate visual hierarchies, and proposed the hierarchical visual perception network for better multiscale learning.
However, lightweight ORSI-SOD is still a desert.
In this paper, we propose an effective and efficient solution for lightweight ORSI-SOD for the first time.
Different from the above lightweight NSI-SOD methods which focus on designing lightweight backbones, we focus on lightening the existing backbone (\ie VGG-16) and proposing effective and lightweight modules.

\subsection{Traditional Methods for ORSI-SOD}
\label{sec:Tra_ORSI_SOD}
Similar to traditional NSI-SOD methods~\cite{2015SODBenchmark}, traditional ORSI-SOD methods also mainly rely on hand-crafted features.
Faur~\etal\cite{2009RDM} presented a rate distortion-based estimation method and considered the mean-shift algorithm to segment the remote sensing images.
Zhang~\etal\cite{2015CIC} applied color information content analysis to ORSIs, and then computed the saliency scores of each color channel and fused color components for final results.
In~\cite{2015SSD}, Zhao~\etal introduced the high-level global and background cues for saliency map integration.
In~\cite{2018SPSS}, Zhang~\etal combined the super-pixel and statistical saliency feature analysis for ORSI-SOD.
Based on low-rank matrix recovery, Zhang~\etal\cite{2019SMFF} proposed a self-adaptive fusion method to fuse color feature, intensity feature, texture, and global contrast for saliency detection in ORSIs.
Huang~\etal\cite{2021CDL} proposed a contrast-weighted dictionary learning based method for VHR RSI saliency detection, which follows the procedure of discriminant dictionary construction, saliency measurement and saliency fusion.

\begin{figure*}
	\centering
	\begin{overpic}[width=1\textwidth]{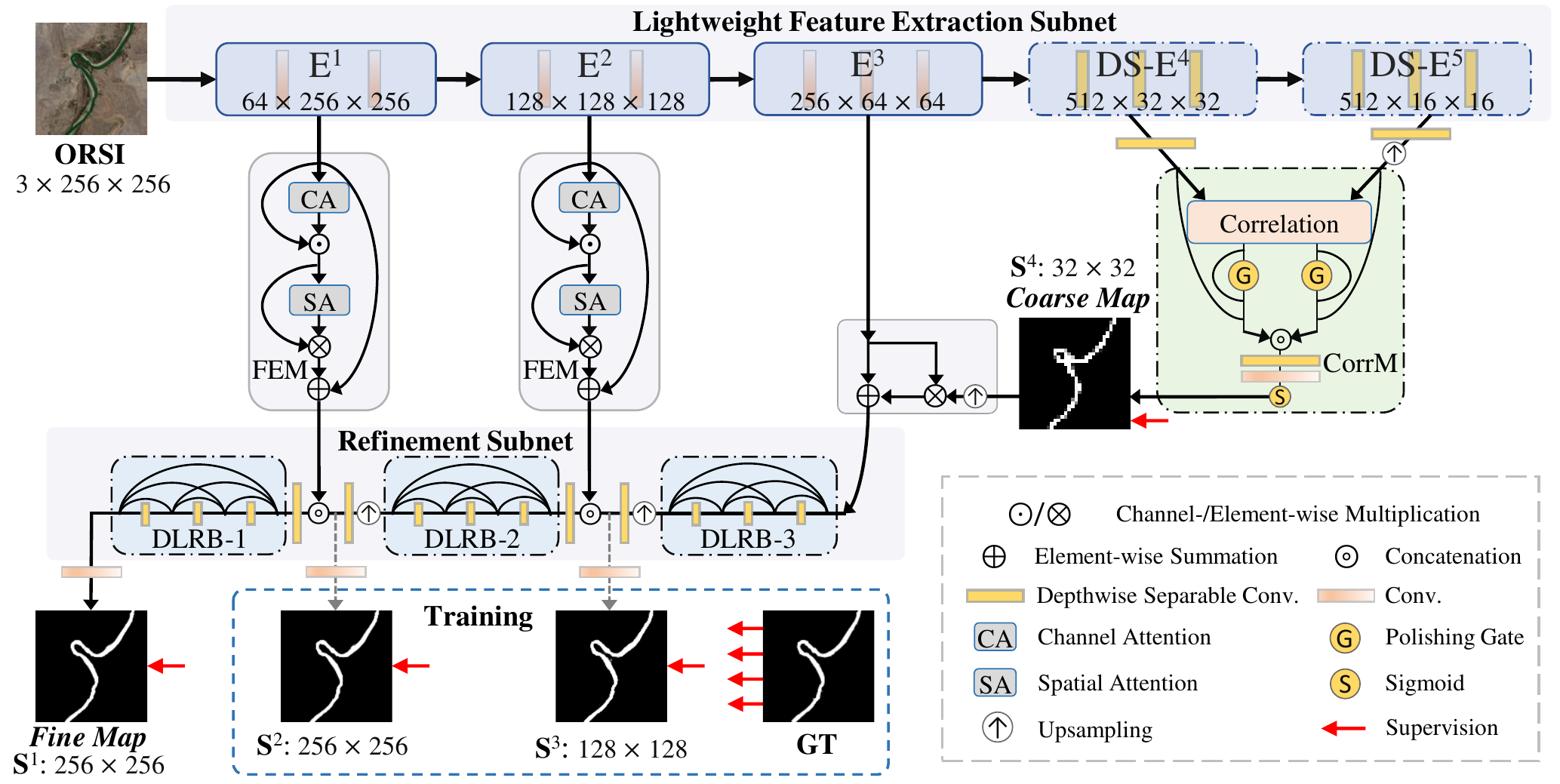}
    \end{overpic}
	\caption{The overall framework of the proposed CorrNet.
	First, we adopt a lightweight feature extraction subnet, which is a variant of classic VGG-16~\cite{2014VGG16ICLR}, to extract the basic feature embeddings.
	Then, we model the cross-layer correlation between two groups of high-level semantic features in the Correlation Module (CorrM), and generate the initial coarse saliency map $\mathbf{S}^{4}$.
	Meanwhile, the low-level detailed features are enhanced in the general Feature Enhancement Module (FEM).
	Finally, we refine the coarse saliency map with the enhanced features in the refinement subnet, which consists of three Dense Lightweight Refinement Blocks (DLRBs), and generate the final fine saliency map $\mathbf{S}^{1}$.
	Notably, in the training phase, we adopt the deep supervision.
    }
    \label{fig:Framework}
\end{figure*}

In addition to general ORSI-SOD methods, there are some traditional methods for specific scenes of ORSIs.
For ship detection, Chen~\etal\cite{2019CR} proposed a contour refinement and the improved generalized Hough transform-based method to handle complex harbor scenes.
For oil tank detection, Liu~\etal\cite{2019CMC} constructed a color Markov chain in the CIE Lab space to generate a bottom-up latent saliency map.
For airport detection, Zhang~\etal\cite{2018VOS} proposed a complementary saliency analysis and saliency-oriented active contour model.
For residential areas extraction, based on complementarities, Zhang~\etal\cite{2016GLSA} merged two global maps and one local map to achieve complete residential areas.

Since hand-crafted features are usually accompanied by a large amount of computational consumption and memory burden, the above traditional methods are not efficient enough for practical applications.

\subsection{CNN-based Methods for ORSI-SOD}
\label{sec:CNN_ORSI_SOD}

Taking the advantage of the powerful feature representation capabilities of CNNs, many CNN-based ORSI-SOD methods have shown good performance.
In order to meet the data requirements of CNN-based methods, Li~\etal\cite{2019LVNet} and Zhang~\etal\cite{2021DAFNet} constructed two challenging datasets for ORSI-SOD, namely ORSSD and EORSSD.
Based on these two datasets, a large number of CNN-based methods have emerged.

In~\cite{2019LVNet}, Li~\etal constructed a L-shape module (\ie the two-stream pyramid module) and a V-shape module (\ie the encoder-decoder module with nested connections) based on features extracted from five different-resolution ORSIs.
Zhou~\etal\cite{2021EMFINet} followed the multi-input strategy, and introduced edge features to complete salient regions in feature level.
Edge information plays an important role in ORSI-SOD.
In~\cite{2021DAFNet}, Zhang~\etal additionally introduced the edge supervision for network training and constructed a multi-task framework.
Tu~\etal\cite{2021MJRBM} extracted edge features from the local cues and the global information, and embed the boundary features into region features.
In addition to edge information, Li~\etal\cite{2021MCCNet} additionally introduced foreground, background and the global image-level content to explore the complementarity of multiple content.
Differently, Zhang~\etal\cite{2021PSL} solved this problem based on the weakly supervised learning.
In~\cite{2020PDFNet}, Li~\etal focused on cross-level feature fusion, and inferred saliency map from a parallel down-up fusion network.
Huang~\etal\cite{2021SARNet} used the high-level features as a guide for locating multi-scale objects, and  combined cross-level features and semantic information to refine the objects.

The above mentioned existing methods have achieved high detection accuracy on the ORSSD and EORSSD datasets.
However, these methods neglect the parameter and computational complexity of models, which prevents them from being deployed into practical systems.
By contrast, in this paper, we no longer focus on improving detection accuracy blindly, but open up a new direction for ORSI-SOD, that is, lightweight ORSI-SOD, which is to achieve a balance among accuracy, parameters, and computational complexity.
To this end, we propose the first lightweight framework, namely CorrNet, for ORSI-SOD.
In CorrNet, we implement all components in a lightweight manner while maintaining competitive or even better performance.

\section{Proposed Method}
\label{sec:OurMethod}
In this section, we elaborate the proposed CorrNet.
In Sec.~\ref{sec:Overview}, we depict the network overview of our CorrNet.
In Sec.~\ref{sec:LFESubnet}, we show how to lighten the backbone.
In Sec.~\ref{sec:CorrM} and Sec.~\ref{sec:DLRB}, we elaborate the Correlation Module and the Dense Lightweight Refinement Block, respectively.
In Sec.~\ref{sec:Loss Function}, we formulate the loss function.


\subsection{Network Overview}
\label{sec:Overview}
We present the overall framework of the proposed CorrNet in Fig.~\ref{fig:Framework}.
CorrNet is comprised of three main components: a lightweight feature extraction subnet (equipped with the general Feature Enhancement Module), a Correlation Module and a refinement subnet (equipped with the Dense Lightweight Refinement Block).
It follows the coarse-to-fine strategy, that is, first generating a coarse saliency map and then sculpting it to generate a fine saliency map.

For feature extraction, we modify the classic vanilla VGG-16~\cite{2014VGG16ICLR} and construct a lightweight feature extraction subnet, named LFE-VGG.
There are five convolution blocks in LFE-VGG.
The first three convolution blocks are denoted as E$^{t}$ ($t=1,2,3$) and their output features as $\boldsymbol{f}^{t}_{\rm e} \in \mathbb{R}^{c_t\!\times\!h_t\!\times\!w_t}$.
The last two convolution blocks are denoted as DS-E$^{t}$ ($t=4,5$), and their output features as $\boldsymbol{f}^{t}_{\rm dse} \in \mathbb{R}^{c_t\!\times\!h_t\!\times\!w_t}$.
The size of input is $\boldsymbol{\mathrm{I}} \in \mathbb{R}^{3\!\times\!256\!\times\!256}$, so $h_t $ is $\frac{256}{2^{t-1}}$, $w_t$ is $\frac{256}{2^{t-1}}$, and $c_t$ belongs to $\{64,128,256,512,512\}$.
Then, we apply the channel and spatial attentions\footnote{Channel attention is implemented by a spatial-wise global max pooling (GMP) and two fully connected layers (the first one is with ReLU and the second one is with sigmoid); and spatial attention is implemented by a channel-wise GMP and a one-channel regular convolution layer with sigmoid.}~\cite{SENet,2018CBAM}
to $\boldsymbol{f}^{1}_{\rm e}$ and $\boldsymbol{f}^{2}_{\rm e}$ in the Feature Enhancement Module (FEM), and get the enhanced features $\boldsymbol{\hat{f}}^{1}_{\rm e}$ and $\boldsymbol{\hat{f}}^{2}_{\rm e}$.
To reduce parameters and computational complexity, we compress the channel of $\boldsymbol{f}^{4}_{\rm dse}$ and $\boldsymbol{f}^{5}_{\rm dse}$ (\ie 512) to 128, and then upsample $\boldsymbol{f}^{5}_{\rm dse}$ to be the same size as $\boldsymbol{f}^{4}_{\rm dse}$.
This way, we get $\boldsymbol{\hat{f}}^{4}_{\rm dse}$ and $\boldsymbol{\hat{f}}^{5}_{\rm dse}$, which belong to $\mathbb{R}^{\hat{c}_4\!\times\!h_4\!\times\!w_4}$ ($\hat{c}_4$ is 128).

Next, we model the feature correlation among $\boldsymbol{\hat{f}}^{4}_{\rm dse}$ and $\boldsymbol{\hat{f}}^{5}_{\rm dse}$ in the Correlation Module (CorrM), aiming to mine the object location information of high-level semantic context.
In this way, we get an initial coarse saliency map $\mathbf{S}^{4}$.
As shown in Fig.~\ref{fig:Framework}, the coarse saliency map $\mathbf{S}^{4}$ can accurately locate the salient objects.
It is used to modulate $\boldsymbol{f}^{3}_{\rm e}$ to focus on the salient regions, generating the modulated features $\boldsymbol{\hat{f}}^{3}_{\rm e}$.
Finally, $\boldsymbol{\hat{f}}^{1}_{\rm e}$, $\boldsymbol{\hat{f}}^{2}_{\rm e}$ and $\boldsymbol{\hat{f}}^{3}_{\rm e}$ are fed to the refinement subnet to generate the final fine saliency map $\mathbf{S}^{1}$ through three Dense Lightweight Refinement Blocks (DLRBs).

\begin{table}[!t]
\centering
\caption{Parameters comparison (including the parameters of convolutional layer and batch normalization layer) of Vanilla-VGG and our LFE-VGG.
  }
\label{LFESubnet}
\renewcommand{\arraystretch}{1.35}
\renewcommand{\tabcolsep}{2.4mm}
\resizebox{0.47\textwidth}{!}{
\begin{tabular}{c||cc|cc}
\bottomrule

 \multirow{2}{*}{Block} & \multicolumn{2}{c|}{Vanilla-VGG} 
 & \multicolumn{2}{c}{LFE-VGG} \\
    &  \#Param(M) & Ratio
    &  \#Param(M) & Ratio \\
\hline
\hline
E$^1$       &  0.04    &   0.27\%      & 0.04   & 1.24\%   \\ 
E$^2$       &  0.22    &   1.50\%      & 0.22   & 6.83\%   \\
E$^3$       &   1.48   &   10.05\%    & 1.48   & 45.96\%   \\
E$^4$/DS-E$^4$       &   \textbf{5.90}   &   \textbf{40.08\%}    &  \textbf{0.67}   & \textbf{20.81\%}   \\
E$^5$/DS-E$^5$       &   \textbf{7.08}   &   \textbf{48.10\%}    &  \textbf{0.81}   & \textbf{25.16\%}   \\
\hline
Total       &   \textbf{14.72}  &   100.00\%    &  \textbf{3.22} & 100.00\%   \\


\toprule
\end{tabular}
}
\end{table}

\subsection{Lightweight Feature Extraction Subnet}
\label{sec:LFESubnet} 
Previous ORSI-SOD methods~\cite{2021MCCNet,2021EMFINet,2021DAFNet} generally adapt the vanilla VGG-16~\cite{2014VGG16ICLR} for basic feature extraction, \ie the last four layers (\ie one max-pooling layer and three fully connected layers) are abandoned.
However, the amount of parameters of the modified vanilla VGG is still large, which is not suitable as the backbone of a lightweight model.
Therefore, in this paper, we propose a convenient way to lighten the vanilla VGG without compromising its feature extraction ability.

In Tab.~\ref{LFESubnet}, we present the amount of parameters (including the parameters of convolution layer and batch normalization layer~\cite{2015BN}) of each convolution block in the vanilla VGG.
We observe that the last two convolution blocks of the vanilla VGG (\ie E$^4$ and E$^5$) have 12.98M parameters, which account for about 88.18\% of all parameters.
Recently, MobileNets~\cite{MobileNet1,MobileNet2} use the depthwise separable convolution (DSConv) to replace the regular convolution.
The DSConv can significantly reduce the parameters without weakening the feature representation ability.
Motivated by~\cite{MobileNet1,MobileNet2}, we adopt DSConvs instead of regular convolutions in E$^4$ and E$^5$, and get the redefined convolution blocks DS-E$^4$ and DS-E$^5$.
There are two reasons why we only redefine E$^4$ and E$^5$.
First, according to the above analysis, E$^4$ and E$^5$ occupy the largest amount of parameters.
Second, we hope to use as many pre-trained parameters of the vanilla VGG as possible to inherit powerful feature extraction ability.

In this way, we construct our Lightweight Feature Extraction Subnet, named LFE-VGG.
As presented in Tab.~\ref{LFESubnet}, the amount of parameters of E$^4$ is reduced from 5.90M to 0.67M, and that of E$^5$ is reduced from 7.08M to 0.81M.
Overall, our LFE-VGG reduces 11.50M parameters compared to the vanilla VGG, and only has 3.22M parameters in total, which is a qualified lightweight backbone.
We will assess the effectiveness and efficiency of LFE-VGG in Sec.~\ref{Ablation Studies}.

\subsection{Correlation Module}
\label{sec:CorrM} 

In video object segmentation, the target objects usually exist in consecutive video frames with small differences.
To accurately segment the target objects, Lu~\etal\cite{2020COSNet} employed the co-attention mechanism~\cite{2016CoAtt} to effectively mine the inherent correlation among consecutive video frames.
Inspired by~\cite{2020COSNet}, considering that the salient regions also exist in consecutive features of an ORSI, we make an attempt to explore the cross-layer correlation of continuous high-level semantic features and propose the Correlation Module.
Different from~\cite{2020COSNet}, which generates corresponding segmentation maps of different video frames, we focus on generating an initial coarse saliency map from the continuous semantic features for location guidance of low-level features.

\begin{figure}
\centering
\footnotesize
  \begin{overpic}[width=0.81\columnwidth]{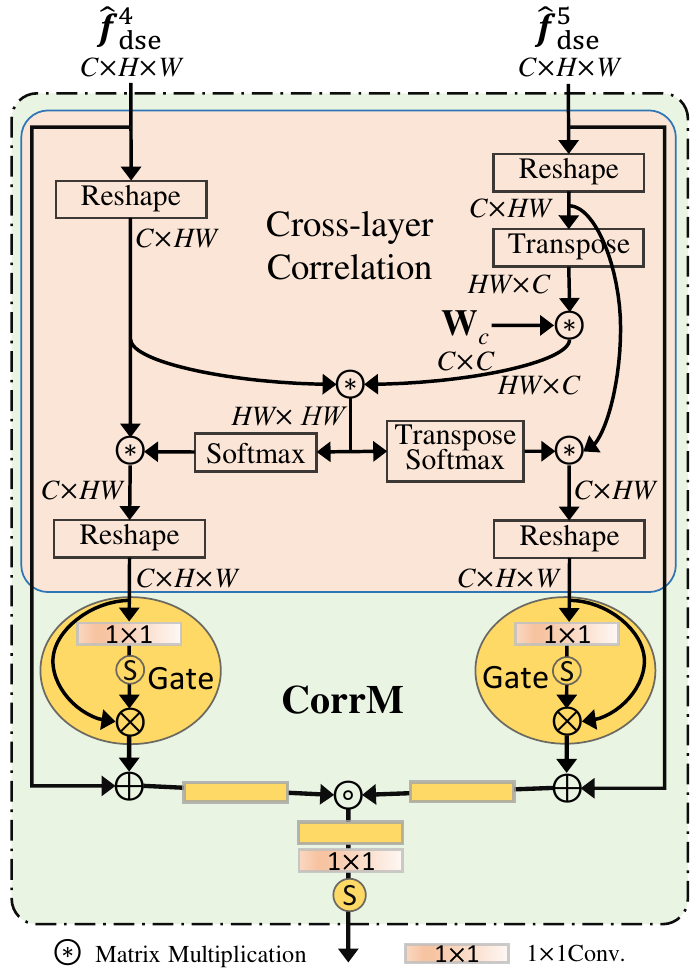}
  \end{overpic}
\caption{
Illustration of the Correlation Module (CorrM).
}
\label{MCCM_structure}
\end{figure}

We illustrate the Correlation Module (CorrM) in Fig.~\ref{MCCM_structure}.
It has three main components: cross-layer correlation operator, polishing gate, and initial coarse saliency map generation.
In the following, we elaborate CorrM based on these three parts, and we also present the feature modulation process based on the coarse saliency map.

\textit{1) Cross-layer Correlation}.
As shown in Fig.~\ref{MCCM_structure}, we perform the cross-layer correlation operator on $\boldsymbol{\hat{f}}^{4}_{\rm dse}$ and $\boldsymbol{\hat{f}}^{5}_{\rm dse}$ (\ie the continuous high-level semantic features), both belonging to $\mathbb{R}^{\hat{c}_4\!\times\!h_4\!\times\!w_4}$.
Here, we simplify refer their sizes ${\hat{c}_4\!\times\!h_4\!\times\!w_4}$ as $C\times\!H\times\!W$ for notation conciseness.

First, we reshape $\boldsymbol{\hat{f}}^{4}_{\rm dse}$ to $\mathbb{R}^{C\times\!(HW)}$, and reshape and transpose $\boldsymbol{\hat{f}}^{5}_{\rm dse}$ to $\mathbb{R}^{(HW)\times\!C}$.
Then, we define a learnable weight matrix $\mathrm{\mathbf{W}}_{c} \in \mathbb{R}^{C\times\!C}$ for $\boldsymbol{\hat{f}}^{5}_{\rm dse}$, and construct a learning process for the cross-layer correlation operator, which makes our CorrM robust.
Next, we compute the feature correlation via matrix multiplication to capture similarity between each row of the reshaped and transposed $\boldsymbol{\hat{f}}^{5}_{\rm dse}$ and each column of the reshaped $\boldsymbol{\hat{f}}^{4}_{\rm dse}$.
We formulate the above process as follows:
\begin{equation}
   \begin{aligned}
    \boldsymbol{r} =  \big({\rm \rm rshp}(\boldsymbol{\hat{f}}^{5}_{\rm dse})\big)^{\top}  \circledast \mathbf{W}_{c} \circledast {\rm rshp}(\boldsymbol{\hat{f}}^{4}_{\rm dse}),
    \label{eq:1}
    \end{aligned}
\end{equation}
where $\boldsymbol{r} \in \mathbb{R}^{(HW)\times\!(HW)}$ is the cross-layer correlation matrix, ${\rm rshp}(\cdot)$ is the reshape operation, ${\top}$ is the matrix transpose operation, and $\circledast$ is the matrix multiplication.

After obtaining the cross-layer correlation matrix $\boldsymbol{r}$, we use the softmax function to normalize it along the rows and columns respectively, and exploit it to determine the location of salient regions of high-level semantic features, which can be formulated as follows:
\begin{equation}
   \begin{aligned}
    \boldsymbol{f}^{4}_{\rm corr} =  {\rm rshp} \big( {\rm rshp}(\boldsymbol{\hat{f}}^{4}_{\rm dse}) \circledast {\rm softmax}(\boldsymbol{r}) \big),
    \label{eq:2}
    \end{aligned}
\end{equation}
\begin{equation}
   \begin{aligned}
    \boldsymbol{f}^{5}_{\rm corr} =  {\rm rshp} \big( {\rm rshp}(\boldsymbol{\hat{f}}^{5}_{\rm dse}) \circledast {\rm softmax}( \boldsymbol{r}^{\top}) \big),
    \label{eq:3}
    \end{aligned}
\end{equation}
where $\{ \boldsymbol{f}^{4}_{\rm corr}, \boldsymbol{f}^{5}_{\rm corr} \} \in \mathbb{R}^{C\!\times\!H\!\times\!W}$ are features containing rich location information.

Since we perform the matrix-based cross-layer correlation operator on $\boldsymbol{\hat{f}}^{4}_{\rm dse}$ and $\boldsymbol{\hat{f}}^{5}_{\rm dse}$, whose sizes are 128$\times$32$\times$32, its computational cost is limited.
Besides, only the parameters of the weight matrix $\mathrm{\mathbf{W}}_{c}$ need to be learned.
Therefore, the cross-layer correlation operator is with few parameters and low computational cost, but has strong capabilities to locate salient objects in ORSIs.

\textit{2) Polishing Gate}.
The above cross-layer correlation operator may leave some redundant information in $\boldsymbol{f}^{4}_{\rm corr}$ and $ \boldsymbol{f}^{5}_{\rm corr} $.
To address this issue, we introduce a simple but effective gate mechanism to polish $\boldsymbol{f}^{4}_{\rm corr}$ and $ \boldsymbol{f}^{5}_{\rm corr} $, and achieve more pure location information.

In order to reduce the module parameters, here, we adopt the regular 1$\times$1 convolution layer to separately produce a response map (which belongs to $[0,1]^{1\!\times\!H\!\times\!W}$) for $\boldsymbol{f}^{4}_{\rm corr}$ and $ \boldsymbol{f}^{5}_{\rm corr} $.
Based on the two response maps, we filter the redundant information of $\boldsymbol{f}^{4}_{\rm corr}$ and $ \boldsymbol{f}^{5}_{\rm corr} $.
We formulate the above gate mechanism as follows:
\begin{equation}
   \begin{aligned}
    \boldsymbol{f}^{4}_{\rm gate} =  {\rm sigmoid} \big( {\rm conv}_{1\times1} (\boldsymbol{f}^{4}_{\rm corr}) \big) \otimes  \boldsymbol{f}^{4}_{\rm corr},\\
    \boldsymbol{f}^{5}_{\rm gate} =  {\rm sigmoid} \big( {\rm conv}_{1\times1} (\boldsymbol{f}^{5}_{\rm corr}) \big) \otimes  \boldsymbol{f}^{5}_{\rm corr},
    \label{eq:5}
    \end{aligned}
\end{equation}
where $\{ \boldsymbol{f}^{4}_{\rm gate}, \boldsymbol{f}^{5}_{\rm gate} \} \in \mathbb{R}^{C\!\times\!H\!\times\!W}$ are the polished features, ${\rm conv}_{1\times1} (\cdot)$ is the regular 1$\times$1 convolution operator, and $\otimes$ is the element-wise multiplication.
Moreover, we adopt the residual connection to merge $\boldsymbol{f}^{4}_{\rm gate}$ and $\boldsymbol{\hat{f}}^{4}_{\rm dse}$, and $\boldsymbol{f}^{5}_{\rm gate}$ and $\boldsymbol{\hat{f}}^{5}_{\rm dse}$, respectively, producing $\boldsymbol{\hat{f}}^{4}_{\rm gate}$ and $\boldsymbol{\hat{f}}^{5}_{\rm gate}$ as follows:
\begin{equation}
   \begin{aligned}
    \boldsymbol{\hat{f}}^{4}_{\rm gate} =  {\rm DSconv} (\boldsymbol{f}^{4}_{\rm gate} \oplus \boldsymbol{\hat{f}}^{4}_{\rm dse} ),\\
    \boldsymbol{\hat{f}}^{5}_{\rm gate} =  {\rm DSconv} (\boldsymbol{f}^{5}_{\rm gate} \oplus \boldsymbol{\hat{f}}^{5}_{\rm dse} ),
    \label{eq:6}
    \end{aligned}
\end{equation}
where ${\rm DSconv}(\cdot)$ is the 3$\times$3 DSConv and $\oplus$ is the element-wise summation.
This original content preservation mode (\ie the residual connection) is good for feature representation.

\textit{3) Initial Coarse Saliency Map Generation}.
Thanks to the above two effective parts of CorrM, the generated $\boldsymbol{\hat{f}}^{4}_{\rm gate}$ and $\boldsymbol{\hat{f}}^{5}_{\rm gate}$ are very informative.
Based on them, we introduce the last part of our CorrM as follows:
\begin{equation}
   \begin{aligned}
       \mathbf{S}^{4} =  {\rm sigmoid} \Big( {\rm conv}_{1\times1} \big( {\rm DSconv}( \boldsymbol{\hat{f}}^{4}_{\rm gate} \circledcirc \boldsymbol{\hat{f}}^{5}_{\rm gate} ) \big) \Big),
    \label{eq:7}
    \end{aligned}
\end{equation}
where $ \mathbf{S}^{4} \in [0,1]^{1\!\times\!32\!\times\!32}$ is the initial coarse saliency map and $\circledcirc$ is the concatenation.
In this way, we completely extract the location information of $\boldsymbol{\hat{f}}^{4}_{\rm gate}$ and $\boldsymbol{\hat{f}}^{5}_{\rm gate}$, accurately determining the salient regions in ORSIs, as $\mathbf{S}^{4}$ shown in Fig.~\ref{fig:Framework}.

\textit{4) Feature Modulation}.
Further, we migrate location information to the basic features $\boldsymbol{f}^{3}_{\rm e}$ as follows:
\begin{equation}
   \begin{aligned}
       \boldsymbol{\hat{f}}^{3}_{\rm e} =  {\rm Up} ( \mathbf{S}^{4} ) \otimes \boldsymbol{f}^{3}_{\rm e} \oplus \boldsymbol{f}^{3}_{\rm e}  ,
    \label{eq:8}
    \end{aligned}
\end{equation}
where $\boldsymbol{\hat{f}}^{3}_{\rm e} \in \mathbb{R}^{c_3\!\times\!h_3\!\times\!w_3}$ is the modulated features and ${\rm Up}(\cdot)$ is the upsampling operation.
This direct feature modulation mode provides accurate location information for the subsequent object refinement process, laying a solid foundation for the final fine saliency map.

In summary, our CorrM balances effectiveness and efficiency, that is, predicting a momentous coarse saliency map with few parameters.
We will evaluate the importance of our CorrM in Sec.~\ref{Ablation Studies}.

\begin{figure}
\centering
\footnotesize
  \begin{overpic}[width=0.93\columnwidth]{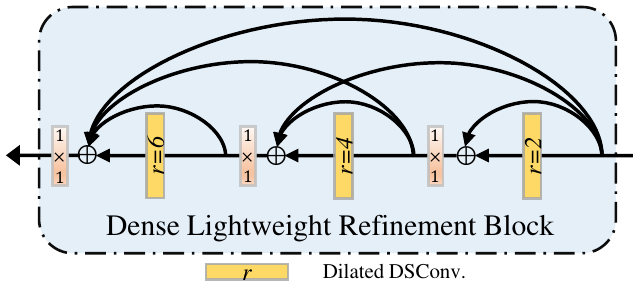}
  \end{overpic}
\caption{
Illustration of the Dense Lightweight Refinement Block (DLRB).
}
\label{DLRB_structure}
\end{figure}

\subsection{Dense Lightweight Refinement Block}
\label{sec:DLRB}
The widely used refinement block usually follows a cascaded structure, \ie several regular convolution layers are connected one by one.
However, there are some challenging scenarios in ORSIs, such as multiple objects and small objects.
The cascaded structure is not conducive to capturing multi-scale information and is a suboptimal way for objects refinement in ORSIs.
Besides, the cascaded refinement block is usually implemented by regular convolution layers, bringing lots of parameters.
Inspired by DenseNet~\cite{2017DenseNet} and DSConv~\cite{MobileNet1,MobileNet2}, we implement a refinement block with the dense structure and DSConvs, constructing a Dense Lightweight Refinement Block (DLRB) for objects refinement in ORSIs, as shown in Fig.~\ref{DLRB_structure}.

For each DLRB, there are three dilated DSConvs with progressive dilation rates \{2,4,6\} and three 1$\times$1 convolution layers.
Dilated DSConvs enlarge the receptive field, capturing multi-scale features comprehensively.
And 1$\times$1 convolution layers are in charge of merging the captured features.
The output feature of DLRB-\emph{t} is denoted as $\boldsymbol{f}^{t}_{\rm dlrb}$.
Here, we take DLRB-3 as an example.
In DLRB-3, its input is $\boldsymbol{\hat{f}}^{3}_{\rm e}$, and we decompose its dense structure into three stages, which are formulated as follows:
\begin{equation}
   \begin{aligned}
       \boldsymbol{f}^{3,1}_{\rm dlrb} =  {\rm conv}_{1\times1} \big({\rm DSconv}_{2} (\boldsymbol{\hat{f}}^{3}_{\rm e}) \oplus \boldsymbol{\hat{f}}^{3}_{\rm e} \big),
    \label{eq:DLRB_1}
    \end{aligned}
\end{equation}
\begin{equation}
   \begin{aligned}
       \boldsymbol{f}^{3,2}_{\rm dlrb} =  {\rm conv}_{1\times1} \big( {\rm DSconv}_{4} (\boldsymbol{f}^{3,1}_{\rm dlrb}) \oplus \boldsymbol{f}^{3,1}_{\rm dlrb} \oplus \boldsymbol{\hat{f}}^{3}_{\rm e} \big),
    \label{eq:DLRB_2}
    \end{aligned}
\end{equation}
\begin{equation}
   \begin{aligned}
       \boldsymbol{f}^{3,3}_{\rm dlrb} =  {\rm conv}_{1\times1} \big( {\rm DSconv}_{6} (\boldsymbol{f}^{3,2}_{\rm dlrb}) \oplus \boldsymbol{f}^{3,2}_{\rm dlrb} \oplus \boldsymbol{f}^{3,1}_{\rm dlrb} \oplus \boldsymbol{\hat{f}}^{3}_{\rm e} \big),
    \label{eq:DLRB_3}
    \end{aligned}
\end{equation}
where ${\rm DSconv}_{r}(\cdot)$ is the dilated 3$\times$3 DSConv with dilation rate $r$, and $\boldsymbol{f}^{3,3}_{\rm dlrb}$ is the output feature of DLRB-3, \ie $\boldsymbol{f}^{3}_{\rm dlrb}$.

In this way, our DLRB can perceive multi-scale information and bring powerful feature representation during the refinement phase, which will facilitate the carving of salient objects in ORSIs and lead to good performance.
We will evaluate the effectiveness of our DLRB in Sec.~\ref{Ablation Studies}.

\begin{table*}[t!]
  \centering
  \small
  \renewcommand{\arraystretch}{1.5}
  \renewcommand{\tabcolsep}{0.8mm}
  \caption{
    Quantitative comparisons with 26 state-of-the-art methods, including five traditional NSI-SOD methods, three traditional ORSI-SOD methods, eleven CNN-based NSI-SOD methods, five CNN-based ORSI-SOD methods, and two lightweight methods, on EORSSD and ORSSD datasets.
   The top three results are highlighted in \textcolor{red}{\textbf{red}}, \textcolor{blue}{\textbf{blue}} and \textcolor{green}{\textbf{green}}, respectively.
    }
\label{table:QuantitativeResults}
  
\resizebox{1\textwidth}{!}{
\begin{tabular}{r|c|c|c|c|c|cccccccc|cccccccc}
\midrule[1pt]    
 \multirow{2}{*}{\normalsize{Methods}}
 & \multirow{2}{*}{\normalsize{Type}}
  & Input
& Speed
& \#Param
& FLOPs
 & \multicolumn{8}{c|}{EORSSD~\cite{2021DAFNet}} 
 & \multicolumn{8}{c}{ORSSD~\cite{2019LVNet}}  \\
 
 \cline{7-14} \cline{15-22} 
       &  &  size & (fps)$\uparrow$ & (M)$\downarrow$ & (G)$\downarrow$ & $S_{\alpha}\uparrow$ & $F_{\beta}^{\rm{max}}\uparrow$ & $F_{\beta}^{\rm{mean}}\uparrow$ & $F_{\beta}^{\rm{adp}}\uparrow$ & $E_{\xi}^{\rm{max}}\uparrow$ & $E_{\xi}^{\rm{mean}}\uparrow$ & $E_{\xi}^{\rm{adp}}\uparrow$ & $ \mathcal{M}\downarrow$
   	          & $S_{\alpha}\uparrow$ & $F_{\beta}^{\rm{max}}\uparrow$ & $F_{\beta}^{\rm{mean}}\uparrow$ & $F_{\beta}^{\rm{adp}}\uparrow$ & $E_{\xi}^{\rm{max}}\uparrow$ & $E_{\xi}^{\rm{mean}}\uparrow$ & $E_{\xi}^{\rm{adp}}\uparrow$ & $ \mathcal{M}\downarrow$\\
	     
\midrule[1pt]
RRWR$_{15}$~\cite{2015RRWR}  & T.N. & - & 0.3 & - & - & .5992 & .3993 & .3686 & .3344 & .6894 & .5943 & .5639 & .1677
								         & .6835 & .5590 & .5125 & .4874 & .7649 & .7017 & .6949 & .1324  \\
HDCT$_{16}$~\cite{2016HDCT}    & T.N. & - & 7    & - & - & .5971 & .5407 & .4018 & .2658 & .7861 & .6376 & .5192 & .1088
									& .6197 & .5257 & .4235 & .3722 & .7719 & .6495 & .6291 & .1309  \\
DSG$_{17}$~\cite{2017DSG}        & T.N. & - & 0.6  & - & - & .6420 & .5232 & .4597 & .4012 & .7260 & .6594 & .6188 & .1246
									& .7195 & .6238 & .5747 & .5657 & .7912 & .7337 & .7532 & .1041  \\
SMD$_{17}$~\cite{2017SMD}        & T.N. & - & -     & - & - & .7101 & .5884 & .5473 & .4081 & .7697 & .7286 & .6416 & .0771
								        & .7640 & .6692 & .6214 & .5568 & .8230 & .7745 & .7682 & .0715  \\
RCRR$_{18}$~\cite{2018RCRR}   & T.N. & - & 0.3 & - & - & .6007 & .3995 & .3685 & .3347 & .6882 & .5946 & .5636 & .1644
									& .6849 & .5591 & .5126 & .4876 & .7651 & .7021 & .6950 & .1277  \\
\hline
VOS$_{18}$~\cite{2018VOS}  	  & T.R. & - & - & - & - & .5082 & .2765 & .2107 & .1836 & .5982 & .4886 & .4767 & .2096
								  & .5366 & .3471 & .2717 & .2633 & .6514 & .5352 & .5826 & .2151 \\
SMFF$_{19}$~\cite{2019SMFF} & T.R. & - & - & - & - & .5401 & .5176 & .2992 & .2083 & .7744 & .5197 & .5014 & .1434
								 & .5312 & .4417 & .2684 & .2496 & .7402 & .4920 & .5676 & .1854 \\
CMC$_{19}$~\cite{2019CMC}  	  & T.R. & - & - & - & - & .5798 & .3268 & .2692 & .2007 & .6803 & 5894 & .4890 & .1057
								 & .6033 & .3913 & .3454 & .3108 & .7064 & .6417 & .5996 & .1267 \\								 
\hline
DSS$_{17}$~\cite{2017DSS}         	& C.N. & 400$\times$300 & 8 & 62.23 & 114.6 & .7868 & .6849 & .5801 & .4597 & .9186 & .7631 & .6933 & .0186 
									 & .8262 & .7467 & .6962 & .6206 & .8860 & .8362 & .8085 & .0363 \\
RADF$_{18}$~\cite{2018RADF}    	& C.N. & 400$\times$400 & 7 & 62.54 & 214.2 & .8179 & .7446 & .6582 & .4933 & .9140 & .8567 & .7162 & .0168 
									 & .8259 & .7619 & .6856 & .5730 & .9130 & .8298 & .7678 & .0382 \\
R3Net$_{18}$~\cite{2018R3Net}   	& C.N. & 300$\times$300 & 2 & 56.16 &  47.5  & .8184 & .7498 & .6302 & .4165 & .9483 & .8294 & .6462 & .0171
									 & .8141 & .7456 & .7383 & .7379 & .8913 & .8681 & .8887 &  .0399\\
PoolNet$_{19}$~\cite{2019PoolNet}  & C.N. & 400$\times$300 & 25 & 53.63 & 123.4 & .8207 & .7545 & .6406 & .4611 & .9292 & .8193 & .6836 & .0210
									    & .8403 & .7706 & .6999 & .6166 & .9343 & .8650 & .8124 & .0358 \\ 
EGNet$_{19}$~\cite{2019EGNet}  	& C.N. & $\sim$380$\times$320 & 9 & 108.07 & 291.9 & .8601 & .7880 & .6967 & .5379 & .9570 & .8775 & .7566 & .0110  
									 & .8721 & .8332 & .7500 & .6452 & .9731 & .9013 & .8226 & .0216 \\ 
GCPA$_{20}$~\cite{2020GCPA}  	& C.N. & 320$\times$320  & 23 & 67.06  & 54.3 & .8869 & .8347 & .7905 & .6723 & .9524 & .9167 & .8647 & .0102  
									 & .9026 & .8687 & .8433 & .7861 & .9509 & .9341 & .9205 & .0168 \\ 
MINet$_{20}$~\cite{2020MINet}  	& C.N. & 320$\times$320 & 12 & 47.56 & 146.3 & .9040 & .8344 & .8174 & .7705 & .9442 & .9346 & .9243 & .0093
									   & .9040 & .8761 & .8574 & .8251 & .9545 & .9454 & .9423 & .0144 \\ 
ITSD$_{20}$~\cite{2020ITSD}  		& C.N. & 288$\times$288 & 16 & 17.08 & 54.5 & .9050 & .8523 & .8221 & .7421 & .9556 & .9407 & .9103 & .0106
									   & .9050 & .8735 & .8502 & .8068 & .9601 & .9482 & .9335 & .0165 \\ 
GateNet$_{20}$~\cite{2020GateNet} & C.N. & 384$\times$384 & 25 & 100.02 & 108.3 & .9114 & .8566 & .8228 & .7109 & .9610 & .9385 & .8909 & .0095
									    & .9186 & .8871 & .8679 & .8229 & .9664 & .9538 & .9428 & .0137 \\ 
SUCA$_{21}$~\cite{2021SUCA}  	& C.N. & 256$\times$256 & 24 & 117.71 & 56.4 & .8988 & .8229 & .7949 & .7260 & .9520 & .9277 & .9082 & .0097
									   & .8989 & .8484 & .8237 & .7748 & .9584 & .9400 & .9194 & .0145 \\
PA-KRN$_{21}$~\cite{2021PAKRN}  & C.N. & 600$\times$600 & 16 & 141.06 & 617.7 & .9192 & .8639 & .8358 & \textcolor{green}{\textbf{.7993}} & .9616 & .9536 & .9416 & .0104
									   & \textcolor{green}{\textbf{.9239}} & .8890 & \textcolor{green}{\textbf{.8727}} & \textcolor{green}{\textbf{.8548}} & .9680 & \textcolor{green}{\textbf{.9620}} & \textcolor{green}{\textbf{.9579}} & .0139 \\									   						
\hline
LVNet$_{19}$~\cite{2019LVNet}  	  & C.R. & 128$\times$128 & 1.4 & -   & - & .8630 & .7794 & .7328 & .6284 & .9254 & .8801 & .8445 & .0146 
									      & .8815 & .8263 & .7995 & .7506 & .9456 & .9259 & .9195 & .0207\\
DAFNet$_{21}$~\cite{2021DAFNet}    & C.R. & 128$\times$128 & 26 & 29.35 & 68.51 & .9166 & .8614 & .7845 & .6427 &  \textcolor{red}{\textbf{.9861}} & .9291 & .8446 & \textcolor{red}{\textbf{.0060}} 
									      & .9191 & \textcolor{green}{\textbf{.8928}} & .8511 & .7876 & \textcolor{blue}{\textbf{.9771}} & .9539 & .9360 & \textcolor{green}{\textbf{.0113}} \\ 
MJRBM$_{21}$~\cite{2021MJRBM} & C.R. & 352$\times$352 & 32 & 43.54 & 95.7 & .9197 & .8656 & .8239 & .7066 & .9646 & .9350 & .8897 & .0099
									   & .9204 & .8842 & .8566 & .8022 & .9623 & .9415 & .9328 & .0163  \\
SARNet$_{21}$~\cite{2021SARNet} & C.R. & 336$\times$336 & 47 & 25.91 & 129.7 & \textcolor{green}{\textbf{.9240}} & \textcolor{green}{\textbf{.8719}} & \textcolor{blue}{\textbf{.8541}} & \textcolor{blue}{\textbf{.8304}} & .9620 & \textcolor{green}{\textbf{.9555}} & \textcolor{blue}{\textbf{.9536}} & .0099
									   & .9134 & .8850 & .8619 & .8512 & .9557 & .9477 & .9464 & .0187  \\
EMFINet$_{21}$~\cite{2021EMFINet} & C.R. & 256$\times$256 & 25 & 107.26  & 480.9 & \textcolor{red}{\textbf{.9290}} & \textcolor{blue}{\textbf{.8720}} & \textcolor{green}{\textbf{.8486}} & .7984 & \textcolor{blue}{\textbf{.9711}} & \textcolor{blue}{\textbf{.9604}} & \textcolor{green}{\textbf{.9501}} & \textcolor{green}{\textbf{.0084}}
									     & \textcolor{blue}{\textbf{.9366}} & \textcolor{blue}{\textbf{.9002}} & \textcolor{blue}{\textbf{.8856}} & \textcolor{blue}{\textbf{.8617}} & \textcolor{green}{\textbf{.9737}} & \textcolor{blue}{\textbf{.9671}} & \textcolor{blue}{\textbf{.9663}} & \textcolor{blue}{\textbf{.0109}}  \\
\hline
CSNet$_{20}$~\cite{2020CSNet} 	& L.W. & 224$\times$224 & 38 & 0.14 & 0.7 & .8364 & .8341 & .7656 & .6319 & .9535 & .8929 & .8339 & .0169
									   & .8910 & .8790 & .8285 & .7615 & .9628 & .9171 & .9068 & .0186 \\
SAMNet$_{21}$~\cite{2021SAMNet} 	& L.W. & 336$\times$336 & 44 & 1.33 & 0.5 & .8622 & .7813 & .7214 & .6114 & .9421 & .8700 & .8284 & .0132
									   & .8761 & .8137 & .7531 & .6843 & .9478 & .8818 & .8656 & .0217 \\

\hline
\hline
\textbf{Ours}				 & L.W. & 256$\times$256 &  100 &  4.09  & 21.09 & \textcolor{blue}{\textbf{.9289}} & \textcolor{red}{\textbf{.8778}} & \textcolor{red}{\textbf{.8620}} & \textcolor{red}{\textbf{.8311}} & \textcolor{green}{\textbf{.9696}} & \textcolor{red}{\textbf{.9646}} & \textcolor{red}{\textbf{.9593}} & \textcolor{blue}{\textbf{.0083}} 
									   & \textcolor{red}{\textbf{.9380}} & \textcolor{red}{\textbf{.9129}} & \textcolor{red}{\textbf{.9002}} & \textcolor{red}{\textbf{.8875}} & \textcolor{red}{\textbf{.9790}} & \textcolor{red}{\textbf{.9746}} & \textcolor{red}{\textbf{.9721}} & \textcolor{red}{\textbf{.0098}}  \\
\toprule[1pt]
\multicolumn{21}{l}{\small{T.N.: Traditional NSI-SOD method, T.R.: Traditional ORSI-SOD method, C.N.: CNN-based NSI-SOD method, C.R.: CNN-based ORSI-SOD method, L.W.: Lightweight method.}} \\
\end{tabular}
}
\end{table*}

\subsection{Loss Function}
\label{sec:Loss Function}
To effectively train CorrNet, we combine the classic BCE loss and IoU loss to construct a comprehensive loss function for network training, which is the same as previous SOD methods~\cite{21HAINet,2019BASNet,2021MCCNet}.
Moreover, we also adopt the deep supervision~\cite{2015DeepSup,2017DSS} in the training phase to supervise two intermediate saliency maps of the refinement subnet as well as the coarse and fine saliency maps, as shown in Fig.~\ref{fig:Framework}.
The intermediate saliency maps and fine saliency map are generated by the 1$\times$1 convolution layer.
We formulate the total loss function ${L}_{\rm total}$ as:
\begin{equation}
   \begin{aligned}
    {L}_{\rm total}  =  \sum\nolimits_{t=1}^4 \Big( \ell_{bce} \big({\rm Up}(\mathbf{S}^{t}),\mathbf{G} \big) + \ell_{iou} \big({\rm Up}(\mathbf{S}^{t}),\mathbf{G} \big) \Big),
    \label{eq:SalLoss}
    \end{aligned}
\end{equation}
where $\mathbf{G}$ is the ground truth, and $\ell_{bce} (\cdot)$ and $\ell_{iou} (\cdot)$ are BCE loss and IoU loss, respectively.

\section{Experiments}
\label{sec:exp}

\subsection{Implementation Details and Evaluation Metrics}
\label{sec:ExpProtocol}

\textit{1) Implementation Details.}
We train and test CorrNet on the ORSSD and EORSSD datasets, respectively.
There are 800 ORSIs with corresponding ground truths in the ORSSD dataset~\cite{2019LVNet}, in which 600 images are used for training and 200 images for testing.
And there are 2,000 ORSIs with corresponding ground truths in the EORSSD dataset~\cite{2021DAFNet}, in which 1,400 images are used for training and 600 images for testing.
We adopt the flipping and rotation for data augmentation, generating 4,800 training pairs for ORSSD and 11,200 training pairs for EORSSD.
All the experiments are conducted on the PyTorch~\cite{PyTorch} platform with an NVIDIA Titan X GPU (12GB memory).
In the training phase, we resized the training pairs to 256$\times$256, and adopt the Adam optimization strategy~\cite{Adam} with the batch size 8 and the initial learning rate $1e^{-4}$, which will be divided by 10 after 30 epochs.
We initialize the first three blocks of LFE-VGG by the pre-trained VGG-16 model~\cite{2014VGG16ICLR}, and initialize other newly added DSConvs and 1$\times$1 convolution layers by the normal distribution~\cite{InitialWei}.
Notably, on the ORSSD dataset, we train our CorrNet for 44 epochs; and on the EORSSD dataset, we train our CorrNet for 34 epochs.

\textit{2) Evaluation Metrics.}
We employ five quantitative evaluation metrics to evaluate our method and other compared methods, including 
S-measure ($S_{\alpha}$, $\alpha$ = 0.5)~\cite{Fan2017Smeasure},
(maximum, mean and adaptive) F-measure ($F_{\beta}$, $\beta^{2}$ = 0.3)~\cite{Fmeasure},
(maximum, mean and adaptive) E-measure ($E_{\xi}$)~\cite{Fan2018Emeasure}, 
Mean Absolute Error (MAE, $\mathcal{M}$), and
Precision-Recall (PR) curve.
The first three evaluation metrics are the bigger the better.
MAE is the smaller the better. 
And PR curve is closer to the upper right, the better.

\subsection{Comparison with State-of-the-art Methods}
We conduct a comprehensive comparison with 26 state-of-the-art NSI-SOD and ORSI-SOD methods, including eight traditional methods (RRWR~\cite{2015RRWR}, HDCT~\cite{2016HDCT}, DSG~\cite{2017DSG}, SMD~\cite{2017SMD}, RCRR~\cite{2018RCRR}, VOS~\cite{2018VOS}, CMC~\cite{2019CMC}, and SMFF~\cite{2019SMFF}),
sixteen CNN-based methods (DSS~\cite{2017DSS}, RADF~\cite{2018RADF}, R3Net~\cite{2018R3Net}, PoolNet~\cite{2019PoolNet}, EGNet~\cite{2019EGNet}, GCPA~\cite{2020GCPA}, MINet~\cite{2020MINet}, ITSD~\cite{2020ITSD}, GateNet~\cite{2020GateNet}, SUCA~\cite{2021SUCA}, PA-KRN~\cite{2021PAKRN},  LVNet~\cite{2019LVNet}, DAFNet~\cite{2021DAFNet}, MJRBM~\cite{2021MJRBM}, SARNet~\cite{2021SARNet}, and EMFINet~\cite{2021EMFINet}), and
two lightweight methods (CSNet~\cite{2020CSNet} and SAMNet~\cite{2021SAMNet}).
Since some methods have different backbone versions, we only report their performance based on VGG backbone.
For a fair comparison, we retain CNN-based NSI-SOD methods with their default parameter settings on the same training set as our method.
And the saliency maps of other methods are provided by the authors or generated by public source codes.

\begin{figure}
\centering
\footnotesize
  \begin{overpic}[width=1\columnwidth]{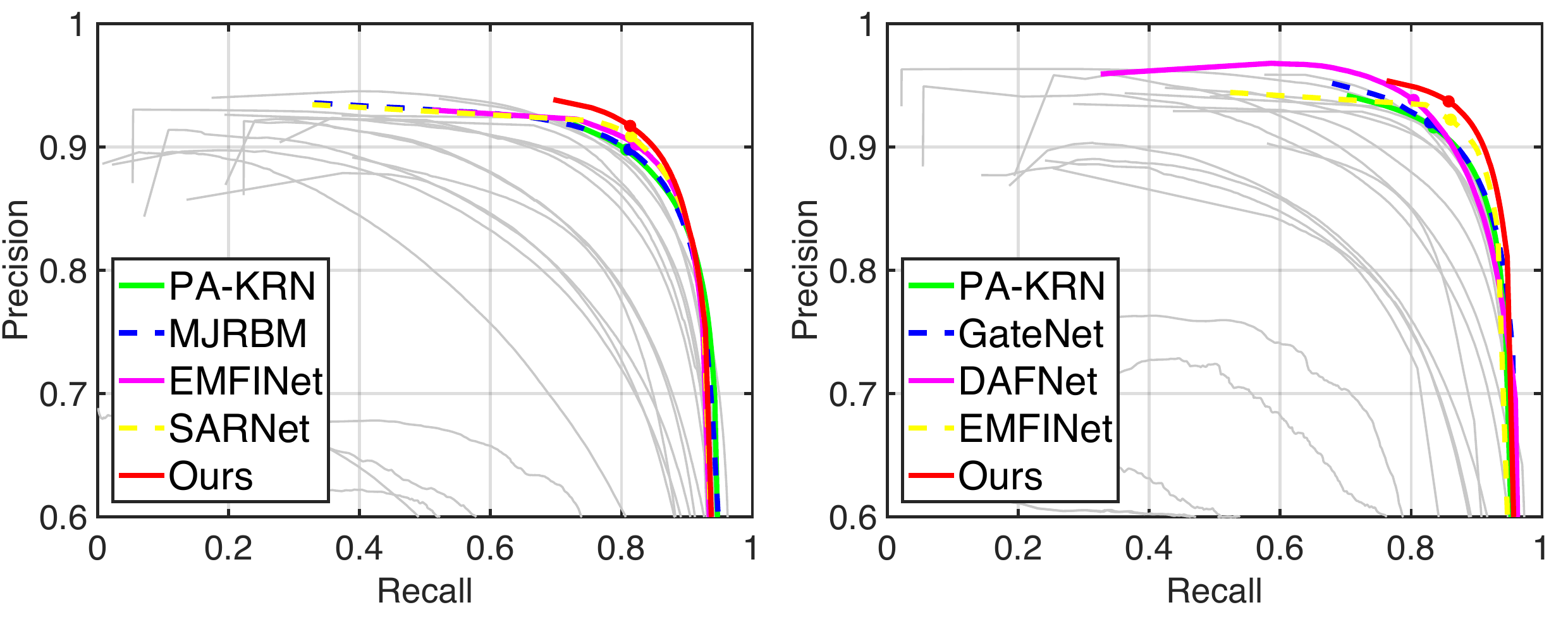}
    \put(15.6,-2.3){ (a) EORSSD~\cite{2021DAFNet} }
    \put(66.6,-2.3){ (b) ORSSD~\cite{2019LVNet} }   
  \end{overpic}
\caption{
Quantitative comparison in terms of PR curve on two datasets.
The top five methods are shown in different colors, while the other compared methods are shown in gray.
Please zoom-in for better visualization of details.
}
\label{PR_comparison}
\end{figure}

\textit{1) Computational Complexity Comparison.}
In Tab.~\ref{table:QuantitativeResults}, we report the inference speed with batch size of 1 (without I/O time), parameter amount (\#Param) and FLOPs of our method and most compared methods.
Notably, our method achieves competitive performance in these three computational complexity metrics.
Our method (\ie 100 fps) has 2.1$\times$ faster inference speed than the second-placed method SARNet (\ie 47 fps).
Compared with CNN-based methods, the parameter amount and FLOPs of our method are smaller than them.
While compared with two lightweight methods, \ie CSNet and SAMNet, our method is slightly inferior, but it is still good as the first lightweight ORSI-SOD solution.
Therefore, we believe that our method is an efficient and promising lightweight ORSI-SOD framework.

\textit{2) Quantitative Comparison.}
In Fig.~\ref{PR_comparison}, we plot the PR curves of our method and all compared methods on EORSSD and ORSSD.
As visible, our method (\ie the \textcolor{red}{red} one) is closest to the upper right than other methods on both datasets, showing a competitive performance.

\begin{figure*}[t!]
    \centering
    \small
	\begin{overpic}[width=1\textwidth]{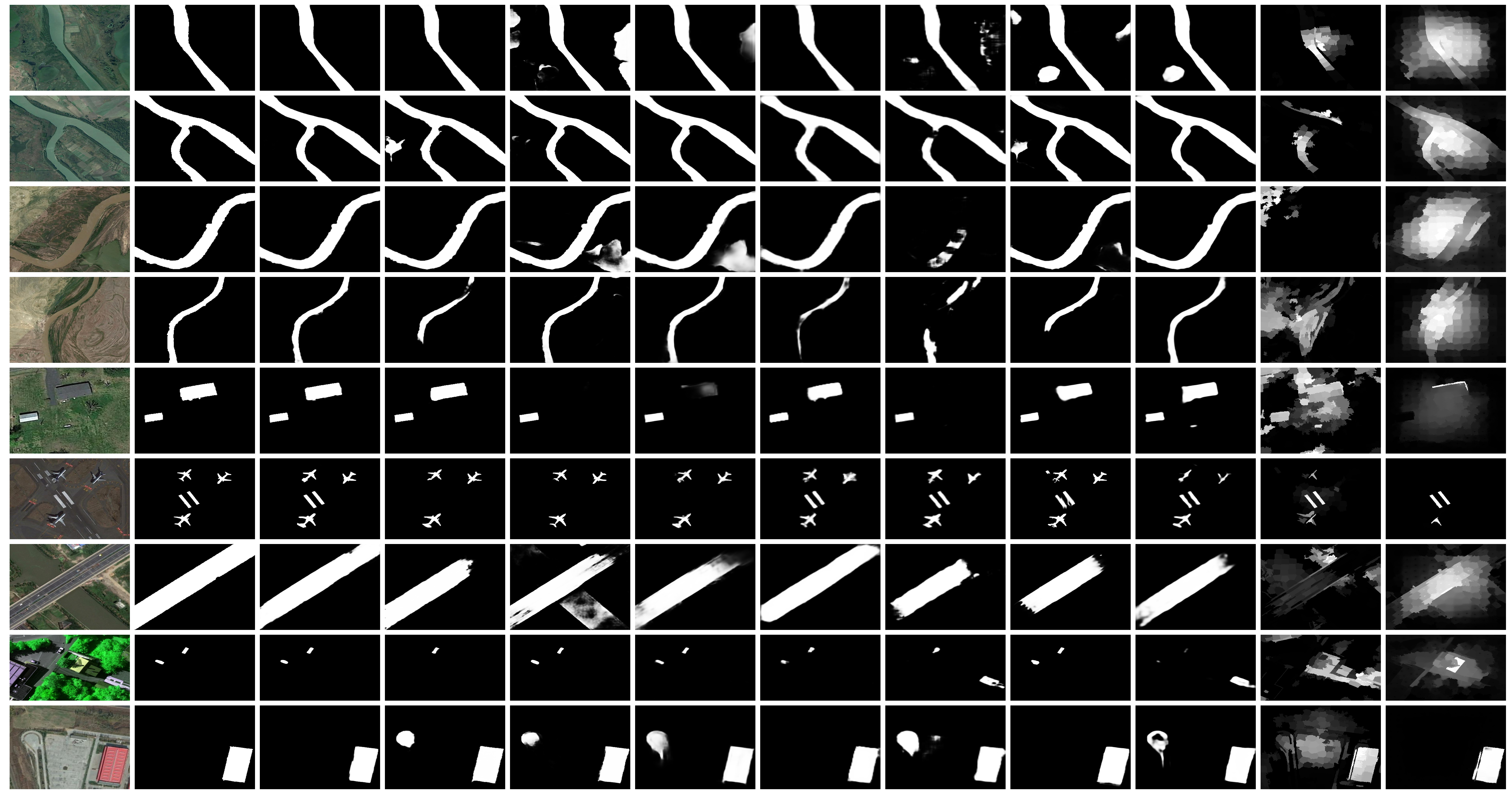}

    \put(2.,-1.35){ ORSI }
    \put(11.13,-1.35){ GT}
    \put(18.62,-1.35){ \textbf{Ours}}
    \put(25.42,-1.35){ EMFINet }
    \put(34.15,-1.35){ SARNet }
    \put(42.3,-1.35){ MJRBM }
    \put(50.67,-1.35){ DAFNet }
    \put(59.65,-1.35){ LVNet }
    \put(67.05,-1.35){ PA-KRN }
    \put(76.2,-1.35){ SUCA }
    \put(84.8,-1.35){ CMC }
    \put(92.7,-1.35){ RCRR } 
    
    \end{overpic}
	\caption{Visual comparisons with nine representative state-of-the-art methods.
	Please zoom-in for the best view.
    }
    \label{fig:VisualExample}
\end{figure*}
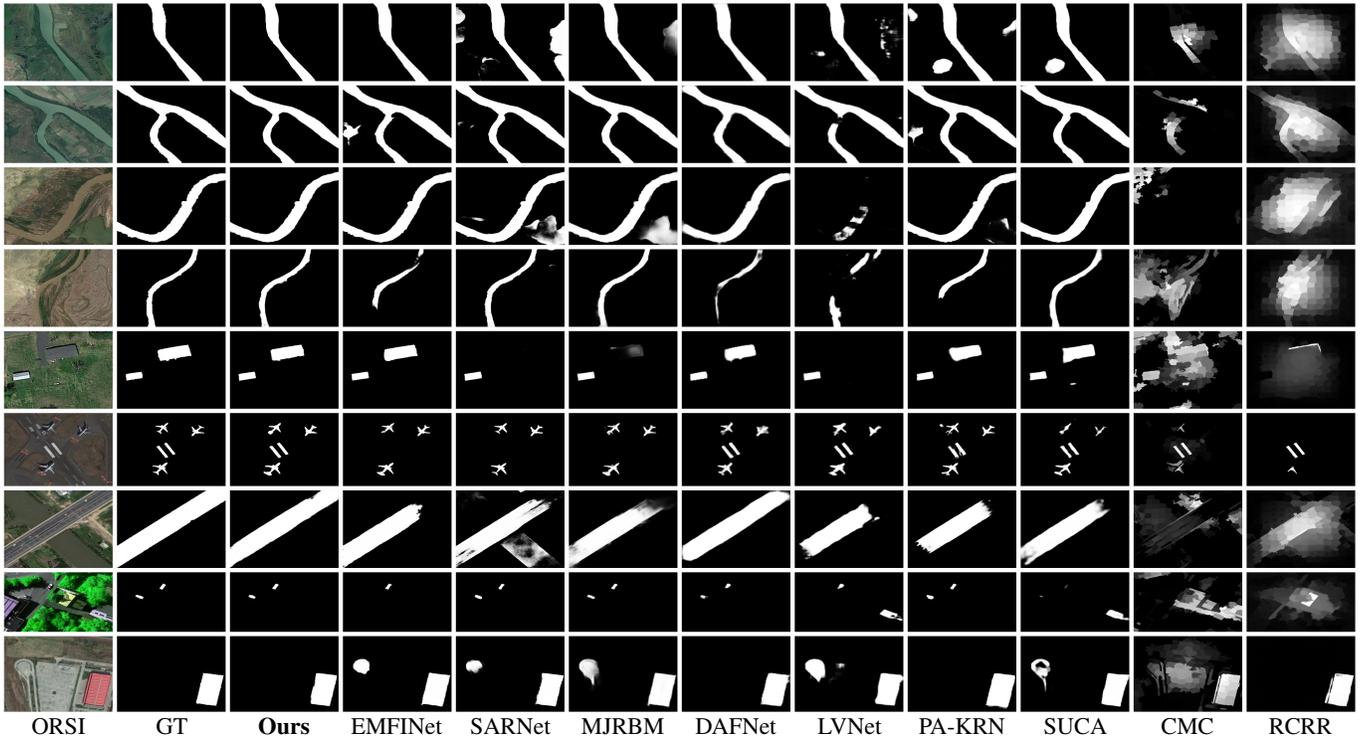

We report the quantitative performance of our method and all compared methods on EORSSD and ORSSD in Tab.~\ref{table:QuantitativeResults}.
On the EORSSD dataset, our method ranks first in five out of all eight metrics.
Compared with EMFINet with similar performance, our method has 4$\times$ faster inference speed, 26$\times$ fewer parameters, and 23$\times$ fewer FLOPs than it.
On the ORSSD dataset, our method outperforms all compared methods in all eight metrics.
Specifically, in $F_{\beta}^{\rm{max}}$, $F_{\beta}^{\rm{mean}}$ and $F_{\beta}^{\rm{apt}}$, our method is 1.27\%, 1.46\% and 2.58\% higher than the second-placed method EMFINet respectively.
Compared with the lightweight methods, \ie CSNet and SAMNet, the performance of our method is significantly better than them.
Moreover, we observe that the performance of CNN-based ORSI-SOD methods is generally better than that of the retrained CNN-based NSI-SOD methods, which indicates that the ORSI scenes are extremely challenging.
Overall, the above analysis clearly demonstrates that our lightweight CorrNet achieves a favorable trade-off between effectiveness and efficiency.

\textit{3) Visual Comparison.}
We present the visual comparison of our method and nine representative state-of-the-art methods on some challenging ORSI scenes in Fig.~\ref{fig:VisualExample}.
As the first four rows of Fig.~\ref{fig:VisualExample}, the first scene is low contrast.
In these four cases, only our method can clearly highlight all salient objects.
Other compared methods are interfered by similar backgrounds and fail in individual cases, such as EMFINet fails in the second case and MJRBM fails in the first and third cases.
As the fifth and sixth rows of Fig.~\ref{fig:VisualExample}, the second scene is multiple objects, which is a difficult scene in ORSI-SOD.
Our method can accurately locate all salient objects with fine details, while some models occasionally miss objects (such as SARNet and LVNet) or fail to outline details (such as DAFNet and SUCA).
As the seventh row of Fig.~\ref{fig:VisualExample}, the third scene is the big object.
In this scene, due to the large span of the bridge, most methods only segment its middle part and ignore its two ends.
As the last two rows of Fig.~\ref{fig:VisualExample}, the fourth scene is cluttered background.
The cluttered background confuses some methods, causing them to incorrectly include background or miss objects in their saliency maps.
Overall, our method shows strong scene adaptability and overcomes the above scenes.

\subsection{Ablation Studies}
\label{Ablation Studies}
Here, we conduct comprehensive experiments to evaluate the effectiveness of important components of our CorrNet on EORSSD dataset.
In particular, we investigate
1) the efficiency of the lightweight feature extraction subnet,
2) the effectiveness of the coarse-to-fine strategy,
3) the individual contribution of each module in CorrNet,
4) the importance of cross-layer correlation and polishing gate in CorrM, and
5) the rationality of dilated DSConvs' dilation rates in DLRB.
For each variant experiment, we rigorously retrain it with the same parameter settings and datasets as in Sec.~\ref{sec:ExpProtocol}.

\begin{table}[!t]
\centering
\caption{Ablation results of evaluating the efficiency of the lightweight feature extraction subnet.
  }
\label{Ablation_backbone}
\renewcommand{\arraystretch}{1.45}
\renewcommand{\tabcolsep}{1.1mm}
\begin{tabular}{c|c|c|cccc}
\bottomrule

 \multirow{2}{*}{Models} & \#Param  & FLOPs 
 & \multicolumn{4}{c}{EORSSD~\cite{2021DAFNet}} \\
 \cline{4-7} 
    &  (M)$\downarrow$ & (G)$\downarrow$
    & $S_{\alpha}\uparrow$ & $F_{\beta}^{\rm{max}}\uparrow$ 
    & $E_{\xi}^{\rm{max}}\uparrow$ & $ \mathcal{M}\downarrow$ \\
\hline
\hline
\emph{Vanilla-VGG}       &  15.59   &   28.1    & .9292 & .8789 & .9718 & .0083   \\ 
\emph{DS-VGG}             &   2.55    &   10.4    & .9063 & .8447 & .9512 & .0108   \\

\hline
LFE-VGG (\textbf{Ours})        &   4.09    &   21.1    & .9289 & .8778 & .9696  & .0083 \\
\toprule
\multicolumn{7}{l}{\footnotesize{\emph{Vanilla-VGG}: VGG-16 with regular convs.}} \\
\multicolumn{7}{l}{\footnotesize{\emph{DS-VGG}: VGG-16 with DSConvs.}} \\
\end{tabular}
\end{table}

\textit{1) The efficiency of the lightweight feature extraction subnet}.
To evaluate the efficiency of the lightweight feature extraction subnet (\ie LFE-VGG), we replace it with two backbones, \ie \emph{Vanilla-VGG} and \emph{DS-VGG}.
\emph{Vanilla-VGG} is all convolution layers of VGG-16 are regular convolution layers, and \emph{DS-VGG} is all convolution layers of VGG-16 are DSConvs.
We report the quantitative results in Tab.~\ref{Ablation_backbone}.

The complete \emph{Vanilla-VGG} does improve performance, but the improvement is limited, \eg $F_{\beta}^{\rm{max}}$ is increased by 0.11\% and $E_{\xi}^{\rm{max}}$ is increased by 0.22\%.
However, along with improved performance comes a massive increase in parameters, \eg \#Param increases sharply from 4.09M to 15.59M.
This means that our LFE-VGG is effective and efficient, and our modification is reasonable.
As for \emph{DS-VGG}, its performance is obviously degraded, \eg $F_{\beta}^{\rm{max}}$ is reduced by 3.31\% and $E_{\xi}^{\rm{max}}$ is reduced by 1.84\%, and its parameter reduction is also limited, \eg \#Param is reduced by 1.54M.
We think that the reason for the performance degradation of \emph{DS-VGG} is because its parameters are initialized by the normal distribution~\cite{InitialWei}, losing the benefits of pre-trained parameters.
Overall, our LFE-VGG maintains the powerful feature extraction ability of E$^1$, E$^2$ and E$^3$, and greatly reduces the parameters of DS-E$^4$ and DS-E$^5$, so it is a qualified lightweight backbone.

\begin{table}[!t]
\centering
\caption{Ablation results of evaluating the effectiveness of the coarse-to-fine strategy.
  The best one in each column is \textbf{bold}.
  }
\label{Ablation_coarse2fine}
\renewcommand{\arraystretch}{1.45}
\renewcommand{\tabcolsep}{2.2mm}
\begin{tabular}{c||cccc}
\bottomrule
 \multirow{2}{*}{Models}  & \multicolumn{4}{c}{EORSSD~\cite{2021DAFNet}}  \\
 \cline{2-5}
    & $S_{\alpha}\uparrow$ & $F_{\beta}^{\rm{max}}\uparrow$ 
    & $E_{\xi}^{\rm{max}}\uparrow$ & $ \mathcal{M}\downarrow$ \\
\hline
\hline

\textbf{S$^4$}              & .8741 & .7800 & .9578  & .0126  \\
\textbf{S$^3$}              & .9138 & .8560 & .9667  & .0094  \\ 
\textbf{S$^2$} 		   & .9265 & .8755 & \textbf{.9698}  & .0083  \\ 
\hline
\textbf{S$^1$} (\textbf{Ours}) & \textbf{.9289} & \textbf{.8778} & .9696  & \textbf{.0083} \\

\toprule
\end{tabular}
\end{table}

\begin{table}[!t]
\centering
\caption{Ablation results of evaluating the individual contribution of each module in CorrNet.
  The best one in each column is \textbf{bold}.
  }
\label{Ablation_component}
\renewcommand{\arraystretch}{1.45}
\renewcommand{\tabcolsep}{0.8mm}
\resizebox{0.49\textwidth}{!}{
\begin{tabular}{c|cccc||cccc}
\bottomrule

 \multirow{2}{*}{No.} & \multirow{2}{*}{Baseline} & \multirow{2}{*}{FEM} & \multirow{2}{*}{DLRB} & \multirow{2}{*}{CorrM}  
 & \multicolumn{4}{c}{EORSSD~\cite{2021DAFNet}}  \\
 
 \cline{6-9}
    & & & & 
    & $S_{\alpha}\uparrow$ & $F_{\beta}^{\rm{max}}\uparrow$ 
    & $E_{\xi}^{\rm{max}}\uparrow$ & $ \mathcal{M}\downarrow$ \\
\hline
\hline
1 &  \Checkmark &                      &                      &                      & .9146 & .8548 & .9535  & .0113  \\
2 &  \Checkmark & \Checkmark  &                      &                      & .9160 & .8591 & .9541  & .0106  \\
3 &  \Checkmark & \Checkmark  & \Checkmark &                      & .9231 & .8679 & .9607  & .0086  \\
4 &  \Checkmark & \Checkmark  &                     & \Checkmark  & .9232 & .8718 & .9639  & .0086  \\ 

\hline
5 &  \Checkmark & \Checkmark  & \Checkmark & \Checkmark  & \textbf{.9289} & \textbf{.8778} & \textbf{.9696}  & \textbf{.0083} \\
\toprule
\end{tabular}
}
\end{table}

\textit{2) The effectiveness of the coarse-to-fine strategy.}
To evaluate the effectiveness of the coarse-to-fine strategy, we quantitatively measure the performance of the initial coarse saliency map ($\mathbf{S}^{4}$), two intermediate saliency maps ($\mathbf{S}^{3}$ and $\mathbf{S}^{2}$) and the final fine saliency map ($\mathbf{S}^{1}$).
As reported in Tab.~\ref{Ablation_coarse2fine}, the quantitative performance is generally incremental, \eg $F_{\beta}^{\rm{max}}:$ 78.00\% $\rightarrow$ 85.60\% $\rightarrow$ 87.55\% $\rightarrow$ 87.78\%.
And compared with $\mathbf{S}^{4}$, the improvement of $\mathbf{S}^{1}$ is greatly significant in all metrics, \ie $S_{\alpha}$, $F_{\beta}^{\rm{max}}$, $E_{\xi}^{\rm{max}}$ and $ \mathcal{M}$ are improved by 5.48\%, 9.78\%, 1.18\% and 0.0043, respectively.
This confirms that the coarse-to-fine manner is useful in our CorrM, and the refinement subnet demonstrates powerful refinement capabilities.

\textit{3) The individual contribution of each module in CorrNet.}
To evaluate the individual contribution of each module, \ie FEM, DLRB and CorrM, we design four variants of the full CorrNet (\ie No.5) in Tab.~\ref{Ablation_component}:
1) \emph{Baseline},
2) \emph{Baseline+FEM},
3) \emph{Baseline+FEM+DLRB}, and
4) \emph{Baseline+FEM+CorrM}.
For \emph{Baseline}, we directly remove FEMs, replace CorrM with concatenation-convolution operation to generate the coarse saliency map, and replace DLRB with three cascaded regular DSConvs.

According to the quantitative performance in Tab.~\ref{Ablation_component}, we observe that each module of CorrNet contributes to the ultimate excellent performance.
Our full CorrNet boosts the primitive \emph{Baseline} by 1.43\%, 2.30\%, 1.61\%, and 0.0030 on $S_{\alpha}$, $F_{\beta}^{\rm{max}}$, $E_{\xi}^{\rm{max}}$ and $ \mathcal{M}$, respectively.
As the key roles of CorrNet, DLRB improves $F_{\beta}^{\rm{max}}$ and $E_{\xi}^{\rm{max}}$ of \emph{Baseline+FEM} by 0.88\% and 0.66\%, respectively, and CorrM improves $F_{\beta}^{\rm{max}}$ and $E_{\xi}^{\rm{max}}$ of \emph{Baseline+FEM} by 1.27\% and 0.98\%, respectively.
With the cooperation of DLRB and CorrM, the full CorrNet improves $F_{\beta}^{\rm{max}}$ and $E_{\xi}^{\rm{max}}$ of \emph{Baseline+FEM} by 1.87\% and 1.55\%, respectively.
Therefore, the above analysis verifies that each module in CorrNet is effective for ORSI-SOD.

\begin{table}[!t]
\centering
\caption{Ablation results of evaluating the importance of cross-layer correlation and polishing gate in CorrM and the rationality of dilated DSconvs in DLRB.
  The best one in each column is \textbf{bold}.
  }
\label{Ablation_variant}
\renewcommand{\arraystretch}{1.45}
\renewcommand{\tabcolsep}{2.2mm}
\begin{tabular}{c||cccc}
\bottomrule
 \multirow{2}{*}{Models}  & \multicolumn{4}{c}{EORSSD~\cite{2021DAFNet}}  \\
 \cline{2-5}
    & $S_{\alpha}\uparrow$ & $F_{\beta}^{\rm{max}}\uparrow$ 
    & $E_{\xi}^{\rm{max}}\uparrow$ & $ \mathcal{M}\downarrow$ \\
\hline
\hline
CorrNet (\textbf{Ours}) & \textbf{.9289} & \textbf{.8778} & \textbf{.9696}  & {.0083} \\

\hline
\textit{w/o Correlation} & .9232 & .8690 & .9649  & .0085  \\ 
\textit{w/o Gate}              & .9253 & .8728 & .9661  & .0086  \\

\hline
\textit{w/o dilation rate}       & .9254 & .8718 & .9663  & .0089  \\
\textit{w/ 1-3-5}               & .9262 & .8727 & .9662  & .0087  \\
\textit{w/ 3-5-7}               & .9272 & .8743 & .9666  & \textbf{.0077}  \\

\toprule
\end{tabular}
\end{table}

\textit{4) The importance of cross-layer correlation and polishing gate in CorrM.}
To evaluate the importance of cross-layer correlation and polishing gate in CorrM, we provide two variants in Tab.~\ref{Ablation_variant}:
1) deleting cross-layer correlation operator in CorrM, namely \textit{w/o Correlation}, and
2) deleting two polishing gates in CorrM, namely \textit{w/o Gate}.
For \textit{w/o Correlation}, there is no cross-layer correlation operator to capture the feature correlation among the high-level semantic context, the object localization capabilities of CorrM are greatly weakened, and $S_{\alpha}$ and $F_{\beta}^{\rm{max}}$ drop to 92.32\% and 86.90\%, respectively.
\textit{w/o Gate} can capture the object location information with some redundant information, and obtain slightly better results than \textit{w/o Correlation}, \ie 92.53\% of $S_{\alpha}$ and 87.28\% of $F_{\beta}^{\rm{max}}$.
The above variants verify that the cross-layer correlation and polishing gate are important to CorrM.

\textit{5) The rationality of dilated DSConvs' dilation rates in DLRB.}
To evaluate the rationality of dilated DSConvs' dilation rates in DLRB, we provide three variants in Tab.~\ref{Ablation_variant}:
1) replacing three dilated DSConvs with three regular DSConvs in DLRB, namely \textit{w/o dilation rate},
2) changing the original dilation rates \{2,4,6\} to \{1,3,5\}, namely \textit{w/ 1-3-5}, and
3) changing the original dilation rates \{2,4,6\} to \{3,5,7\}, namely \textit{w/ 3-5-7}.
Based on the quantitative performance of \textit{w/o dilation rate} and \textit{w/ 1-3-5}, we observe that the relatively large receptive fields can capture more complementary multi-scale information, which is very important for objects refinement in ORSI-SOD.
However, when we continue to expand the receptive fields to \{3,5,7\}, we observe the performance degradation of \textit{w/ 3-5-7}, possibly due to that excessively large receptive fields make DLRB impossible to accurately capture the salient objects with variable scales in ORSIs.
Thus, we can come to a conclusion that the dilated DSConvs with dilation rates \{2,4,6\} are rational and exactly appropriate in DLRB.

\subsection{Discussion}
\label{Discussion}


Here, we discuss the weaknesses of our method and our future works.
For weaknesses, we summarize as follows:
1) since our method is based on GPU, its model size is still too large for edge computing devices; and
2) although the parameter amount and computations of our method are small as compared with most CNN-based methods, it is still difficult to run on the CPU in real time.

Therefore, in future works, we will work in the following two directions:
1) similar to the lightweight ORSI-SOD methods, we will focus on developing a lightweight backbone with smaller model size for ORSI-SOD; and
2) we will introduce the model pruning technology into our model to remove redundant layers and to further accelerate our model.

\section{Conclusion}
\label{sec:con}
In this paper, we propose an effective lightweight framework, namely CorrNet, for ORSI-SOD.
In CorrNet, we first lighten the vanilla VGG and propose a lightweight feature extraction subnet, namely LFE-VGG.
Then, we lighten other modules of CorrNet, that is, we use DSConvs instead of regular convolution layers in the modules.
In addition, in order to obtain a good performance, CorrNet follows the coarse-to-fine strategy.
It first generates an initial coarse saliency map from high-level semantic features via the Correlation Module, and then refines the salient objects via the refinement subnet equipped with Dense Lightweight Refinement Blocks, producing the final fine saliency map.
Experimental evaluations on two ORSI-SOD datasets demonstrate that though our CorrNet only has 4.09M parameters and 21.09G FLOPs, it achieves competitive or even better performance than large CNN-based methods and runs at 100 fps.
The success of our CorrNet comes from three aspects: 1) the matrix-based cross-layer correlation operation that extracts salient regions effectively and only contains a few parameters, 2) the DSConv that maintains powerful feature representation capabilities and only has about 10\% of the parameters of the regular convolution layer, and 3) the coarse-to-fine strategy that lays a high-accuracy foundation.



\ifCLASSOPTIONcaptionsoff
  \newpage
\fi

\bibliographystyle{IEEEtran}
\bibliography{ORSIref}

%



%

\end{document}